%% file: main.tex
\documentclass[runningheads]{llncs}

\usepackage[]{eccv}

\usepackage{eccvabbrv}

\usepackage{graphicx}
\usepackage{booktabs}

\usepackage[accsupp]{axessibility}  

\usepackage{hyperref}
\usepackage{multirow}
\usepackage{tabularx}
\usepackage{float}
\usepackage{makecell}
\usepackage{colortbl}
\usepackage{pgffor}
\usepackage{graphicx}
\usepackage{subcaption}
\usepackage{float}
\usepackage{placeins}

\usepackage{orcidlink}

\newcommand{\equalcontrib}{\textsuperscript{*}}
\newcommand{\corrauthor}{\textsuperscript{\textdagger}}

\begin{document}
\emergencystretch=2em

\title{WAPR: A Foundation Model for Wide-Angle Refinement in Unseen Object Pose Estimation} 

\titlerunning{WAPR for Wide-Angle Pose Refinement}

\author{
	Yulin Wang\equalcontrib\inst{1}\orcidlink{0009-0000-4625-2527}
	\and
	Mengting Hu\equalcontrib\inst{1}\orcidlink{0009-0002-2897-5996}
	\and
	Hongli Li\inst{2}\orcidlink{0009-0001-7443-5563}
	\and
	Jianghao Zhou\inst{1}\orcidlink{0009-0009-9392-6665}
	\and
	Chen~LUO\corrauthor\inst{1}\orcidlink{0000-0003-4339-0355}
}

\authorrunning{Y. Wang et al.}

\institute{
	Southeast University, Nanjing, China\\
	\email{yulinwang@seu.edu.cn, 220240361@seu.edu.cn,}\\
	\email{zhou\_jianghao@seu.edu.cn, chenluo@seu.edu.cn}
	\and
	Purdue University, West Lafayette, IN, USA\\
	\email{li5125@purdue.edu}\\
	\equalcontrib Equal contribution.
	\quad
	\corrauthor Corresponding author.
}

\maketitle

\begin{abstract} 
Real-world applications require 6D pose estimation to be accurate, fast, and scalable to unseen objects. This paper introduces WAPR, a zero-shot wide-angle pose refinement model that refines candidate poses with rotational deviations up to \(90^\circ\). With as few as 12 candidate poses per detected object instance, WAPR supports fast inference within \(1\) s per frame and reaches a pose-estimation throughput of up to 25 detected object instances per second. {\color{black} To support wide-angle training for rotationally symmetric objects, WAPR uses rotational symmetry priors to canonicalize symmetry-equivalent pose targets before loss computation. We further construct SA6D, a large-scale 6D training dataset with such priors. SA6D obtains KASAL-assisted rotational symmetry priors for 944 GSO scans and expands them through geometry and texture augmentation into \(\sim\)50K augmented object instances and \(\sim\)2M rendered RGB-D images.} In addition, an angle-balanced loss stabilizes learning across different angular ranges by reducing the influence of uninformative large-error cases. Experiments on seven BOP core datasets show that WAPR achieves state-of-the-art performance in unseen-object 6D pose localization and detection under both fast and unconstrained inference settings. {\color{black} Project page: \url{https://github.com/WangYuLin-SEU/WAPR}.}


\end{abstract}

\input{sec/1_intro}
\input{sec/2_relatedwork}
\input{sec/3_methodology}

\input{sec/4_experiment}
\FloatBarrier
\input{sec/5_conclusion}
\input{sec/Acknowledgments}
%
%
\clearpage
{
	\nocite{he_resnet_2015}
	\bibliographystyle{splncs04}
	\bibliography{main}
}

\end{document}

%% file: sec/1_intro.tex
\section{Introduction}
\label{sec:intro}
{\color{black} 6D object pose estimation \cite{ferrari_bop_tud-l_2018,hodan_bop_metric_2020,hodan_bop_2024,nguyen_bop_2025} is a fundamental problem in computer vision, with wide-ranging applications in augmented reality \cite{park_ar1_2024,li_ar2_2024,yang_ar3_2023}, robotics \cite{liu_robotic1_2023,yu_robotic2_2024,zhang_robotic3_2022,thalhammer_robotic4_2024}, autonomous driving \cite{hoque_ad1_2023,zou_ad2_2021,wu_ad3_2019}, and industrial automation \cite{bib1,bib2}. Traditional pipelines typically combine 2D detection or segmentation \cite{ge_yolox_2021,he_maskrcnn_2017,ren_fast-rcnn_2017,tian_fcos_2019} with 6D pose estimation \cite{HccePose_BF,wang_gdr-net:_2021,su_zebrapose:_2022}, followed by optional refinement \cite{zakharov_dpod:_2019,li_deepim:_2020}. While effective for trained objects, their reliance on object-specific training limits scalability in real-world deployments.}

\begin{figure}[t]
	\centering
	\includegraphics[width=0.8\linewidth]{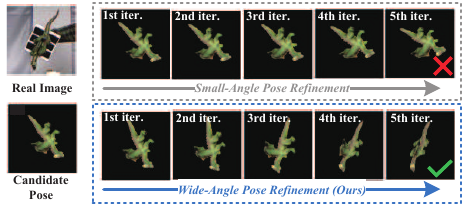}
	\caption{Visualization of multi-iteration refinement for candidate poses with large angular deviations. The example is from TUD-L \cite{ferrari_bop_tud-l_2018}, where the handheld object moves continuously. With large deviations between the candidate pose and the ground-truth pose, conventional small-angle refiners often fail, whereas WAPR corrects the pose through multi-iteration refinement.}
	\label{fig1:onecol}
\end{figure}

{\color{black} To improve scalability, unseen-object pose estimation \cite{caraffa_freeze_2024,wen_foundationpose:_2024,Co-op} estimates poses for novel objects from object models with minimal or no retraining. Recent methods \cite{caraffa_freeze_2024,moon_genflow_2024,leonardis_foundpose_2025,Co-op,nguyen2024gigaPose} commonly follow a candidate-based pipeline, where coarse pose hypotheses are generated from rendered views or feature correspondences and then refined or scored. This design makes candidate coverage and refinement range closely coupled. Using many candidates improves pose coverage but increases runtime, while sparse candidates are efficient but may leave larger angular deviations to the refiner. As shown in Fig. \ref{fig1:onecol}, such deviations can arise from coarse hypotheses, occlusion, and large viewpoint changes, motivating a refinement model that remains reliable across a wider angular range.}

{\color{black} This paper introduces WAPR, a zero-shot foundation model for wide-angle pose refinement. WAPR follows the render-and-compare paradigm by comparing a rendered RGB-D observation and mask under a candidate pose with the real RGB-D input, and directly regresses rotation and translation updates. During training, ground-truth poses are perturbed by up to \(90^\circ\), allowing WAPR to learn pose updates over a wide angular range. At inference time, this wide-angle refinement capability enables accurate pose estimation with a small number of candidates, including only 12 candidates in the fast setting.}

{\color{black} Wide-angle training also makes pose ambiguity more prominent for rotationally symmetric objects. Different symmetry-equivalent poses may produce indistinguishable or nearly indistinguishable observations, and assigning arbitrary raw rotation targets can make visually similar candidate--target pairs receive inconsistent supervision. Following the common practice in 6D pose estimation of treating symmetry-equivalent poses consistently \cite{xiang_posecnn_ycb-v_2018,CosyPose,su_zebrapose:_2022,HccePose_BF}, WAPR uses rotational symmetry priors to canonicalize symmetry-equivalent pose targets before loss computation. This makes wide-angle candidate--target pairs well-defined during training.}

{\color{black} To support this training at scale, this paper constructs SA6D, a large-scale 6D dataset with rotational symmetry priors. Large-scale synthetic datasets for unseen-object pose estimation \cite{labbe_megapose_2022,wen_foundationpose:_2024} provide diverse object models and rendered RGB-D supervision. SA6D complements this direction by adding dataset-level rotational symmetry priors. It uses KASAL \cite{KASAL} to obtain reliable symmetry priors for 944 GSO \cite{gso} scans, and expands these annotated assets through geometry and texture augmentation into \(\sim\)50K augmented object instances and 2M synthetic RGB-D images. The training objective also includes an angle-balanced loss, which reduces the dominance of low-correspondence large-error samples during wide-angle optimization.}

{\color{black} Experiments on seven core BOP datasets \cite{fleet_lmo_2014,hodan_t-less_2017,drost_itodd_2017,kaskman_hb_2019,xiang_posecnn_ycb-v_2018,doumanoglou_ic-bin_2016,ferrari_bop_tud-l_2018} evaluate WAPR for zero-shot unseen-object 6D pose localization and detection. In the fast setting (\(<1\) s per frame) with MUSE detections, WAPR reaches 80.6\% AR / 78.6\% AP, improving over Co-op \cite{Co-op} by +4.7\% AR / +9.3\% AP. With Co-op's F3DT2D detector under the same fast-inference protocol, WAPR obtains 78.1\% AR / 76.5\% AP, giving a matched-detector gain of +2.2\% AR / +7.2\% AP. In the unconstrained multi-detector setting, WAPR reaches 84.5\% AR / 85.3\% AP, improving over FreeZeV2 \cite{caraffa_freeze_2024,freezev2} by +1.2\% AR / +2.0\% AP.}

{\color{black} The main contributions are as follows.}
\begin{itemize}
	\item {\color{black} This paper presents WAPR, a zero-shot foundation model for wide-angle pose refinement. With render-and-compare RGB-D inputs and iterative refinement, WAPR handles rotational deviations up to \(90^\circ\) and enables accurate pose estimation with sparse candidate poses.}
	\item {\color{black} This paper constructs SA6D, a large-scale 6D dataset with rotational symmetry priors. SA6D uses a cost-efficient KASAL-assisted strategy to expand 944 annotated GSO scans into \(\sim\)50K augmented object instances and 2M synthetic RGB-D images.}
	\item {\color{black} This paper develops wide-angle training and sparse candidate-based inference for unseen-object pose estimation. Rotational symmetry priors canonicalize ambiguous training pairs, angle-balanced loss stabilizes wide-angle learning, and score-based selection produces accurate final poses from a small candidate set.}
\end{itemize}


%% file: sec/2_relatedwork.tex
\section{Related Works}

{\color{black} This section reviews 6D pose estimation for seen objects \cite{liu_gdrnpp_2022,su_zebrapose:_2022,hong_rdpn6d_2024,hai_radet_2023,wu_kcv_2022,haugaard_surfemb_2022}, unseen-object 6D pose estimation \cite{caraffa_freeze_2024,Co-op,wen_foundationpose:_2024,lin_sam-6d:_2024,moon_genflow_2024,labbe_megapose_2022}, and symmetry handling in 6D pose learning and evaluation.}

\subsection{6D Pose Estimation for Seen Objects}
{\color{black} Seen-object 6D pose estimation assumes that training data are available for each target object. Early methods directly regress object rotation and translation from RGB or RGB-D observations \cite{xiang_posecnn_ycb-v_2018,bib3}. Later methods improve robustness by predicting keypoints or dense surface coordinates and recovering the pose with PnP \cite{HccePose_BF,peng_pvnet:_2018,hodan_epos:_2020,bib4,bib5,epnp}. This correspondence-based formulation provides explicit 2D--3D constraints and is effective under partial occlusion.}

{\color{black} Pose refinement further improves accuracy by aligning rendered observations with real images. Representative methods such as DeepIM \cite{li_deepim:_2020} and DPOD \cite{zakharov_dpod:_2019} refine an initial pose by comparing rendered and observed images. Recent methods such as GDR-Net \cite{wang_gdr-net:_2021} and ZebraPose \cite{su_zebrapose:_2022} combine detection, dense correspondence prediction, PnP, and optional refinement to achieve strong performance on trained objects. These object-specific pipelines provide the basis for many recent unseen-object methods, while scalable deployment to newly introduced objects motivates methods that reduce object-specific training.}

\subsection{6D Pose Estimation for Unseen Objects}
{\color{black} Unseen-object 6D pose estimation estimates poses for novel objects from object models with minimal or no object-specific retraining. MegaPose \cite{labbe_megapose_2022} trains large-scale render-and-compare models on synthetic data and demonstrates strong generalization to unseen objects. FoundationPose \cite{wen_foundationpose:_2024} combines pose initialization, iterative refinement, and pose scoring in a unified framework. SAM-6D \cite{lin_sam-6d:_2024} and CNOS \cite{nguyen_cnos_2023} use large visual models for zero-shot object localization and segmentation. Recent methods including FreeZe \cite{caraffa_freeze_2024}, GenFlow \cite{moon_genflow_2024}, FoundPose \cite{leonardis_foundpose_2025}, GigaPose \cite{nguyen2024gigaPose}, and Co-op \cite{Co-op} improve different stages of the pipeline, including object detection, coarse pose retrieval, feature matching, refinement, and pose selection.}

{\color{black} Most unseen-object pipelines generate candidate poses from rendered templates or correspondence matching, refine these candidates, and then select the final pose. This design makes candidate coverage and inference efficiency closely connected. Dense candidate sampling improves pose coverage but increases runtime, while sparse candidates are more efficient but place higher demand on the refinement stage. Many existing refinement modules are most effective when the initial hypothesis is already close to the target pose. WAPR targets the complementary wide-angle regime, where sparse candidates leave larger residual deviations for the refiner.}

\subsection{Symmetry Handling in 6D Pose Learning and Evaluation}
{\color{black} Object symmetry is a long-standing issue in 6D pose estimation because multiple poses may correspond to the same or nearly the same visual observation. Benchmarks and evaluation protocols therefore use symmetry-aware pose metrics to avoid penalizing equivalent poses \cite{hodan_bop_metric_2020}. Learning-based methods also incorporate symmetry into supervision. PoseCNN \cite{xiang_posecnn_ycb-v_2018} introduces the ShapeMatch loss to make pose learning invariant to indistinguishable symmetric poses, while CosyPose \cite{CosyPose} uses symmetry-aware pose objectives for object-specific pose estimation. Coordinate-based methods such as ZebraPose \cite{su_zebrapose:_2022} and HccePose \cite{HccePose_BF} also rely on symmetry-consistent definitions for learning or evaluation.}

{\color{black} These works establish symmetry-consistent supervision as a standard principle in 6D pose estimation. Beyond loss design and evaluation, rotational symmetry priors are also useful as dataset-level annotations when synthetic pose pairs are generated for training. For wide-angle refinement, candidate and target poses may span symmetry-equivalent pose regions, and such pairs should receive consistent pose-update targets. Large-scale synthetic datasets have mainly focused on object diversity, rendering realism, and scalable pose supervision \cite{labbe_megapose_2022,wen_foundationpose:_2024}. SA6D complements this direction by adding KASAL-assisted rotational symmetry priors at scale and using them to canonicalize synthetic candidate--target pose pairs before training.}

%% file: sec/3_methodology.tex
\section{Methodology}

\begin{figure}[t]
	\centering
	\includegraphics[width=0.90\linewidth]{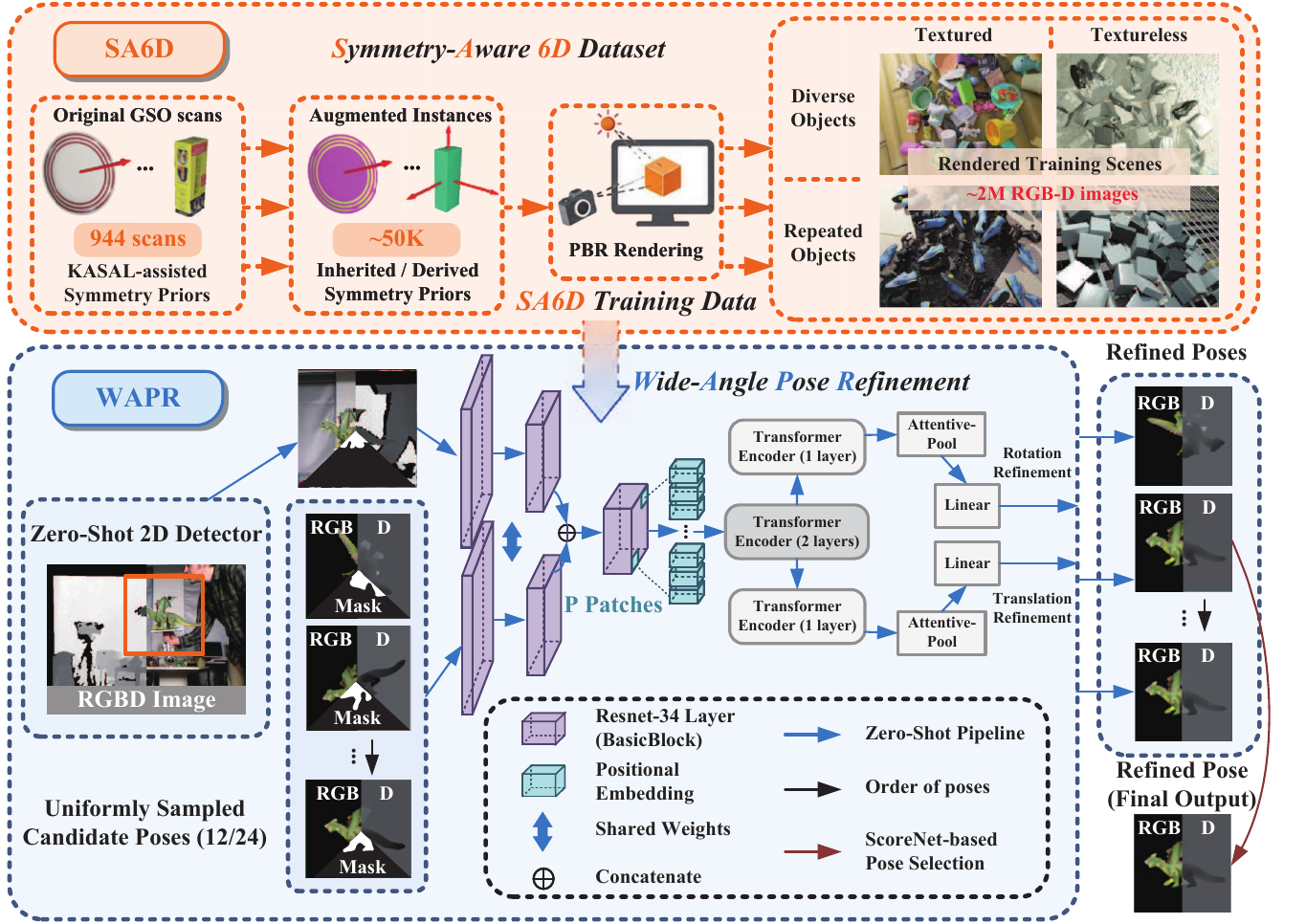}
	\hfill
	\caption{{\color{black} Overview of the proposed framework. Top: SA6D constructs large-scale RGB-D training data with KASAL-assisted rotational symmetry priors. These priors are used to canonicalize symmetry-equivalent training pairs. Bottom: WAPR refines sparse candidate poses by comparing real and rendered RGB-D observations with masks. The refined candidates are selected by a score-based pose selection module, enabling accurate 6D pose estimation under large initial misalignments of up to \(90^\circ\).}}
	\label{fig2:twocol}
\end{figure}

{\color{black} Figure \ref{fig2:twocol} summarizes the proposed framework. It consists of two connected parts. First, SA6D provides large-scale synthetic RGB-D training data with rotational symmetry priors. These priors are used to canonicalize symmetry-equivalent candidate--target pose pairs before training, so that wide-angle pose updates are well-defined for rotationally symmetric objects. Second, WAPR learns a render-and-compare refinement model that predicts rotation and translation updates from paired real/rendered RGB-D observations and masks. An angle-balanced loss further stabilizes training over a wide range of pose deviations, and sparse candidate generation enables efficient inference with only 12 or 24 candidates.}

\begin{figure}[t]
	\centering
	\includegraphics[width=0.8\linewidth]{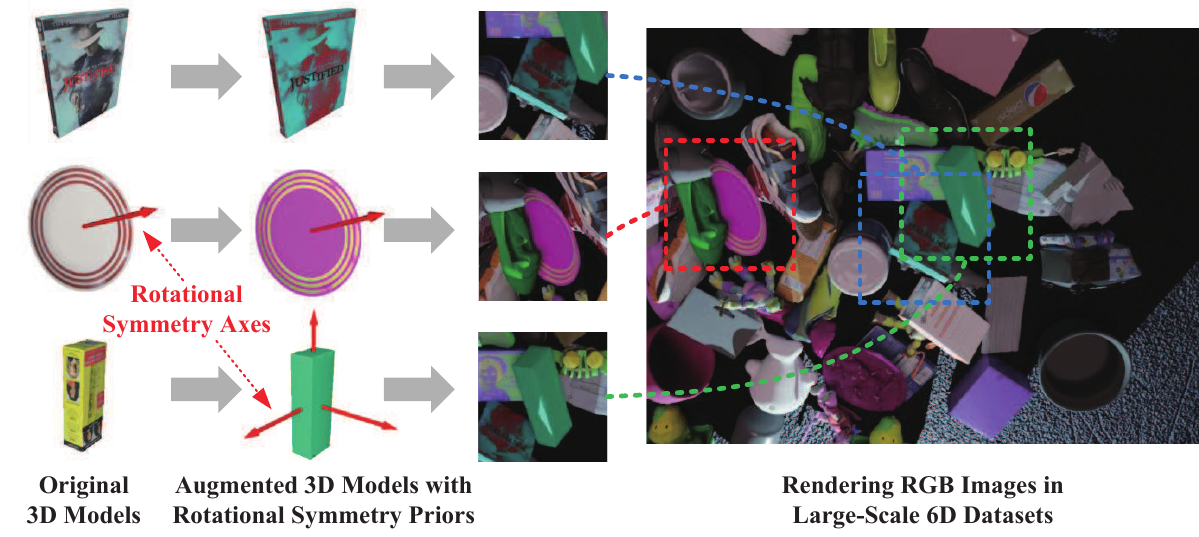}
	\caption{{\color{black} Large-scale construction of SA6D with rotational symmetry priors. KASAL-assisted annotation provides texture-aware and geometry-only priors for the original GSO scans. Geometry scaling and texture randomization reuse these priors to generate \(\sim\)20K non-symmetric and \(\sim\)30K rotationally symmetric augmented instances, and BlenderProc \cite{denninger_blenderproc_2019} renders \(\sim\)2M synthetic RGB-D images for training.}}
	\label{fig4}
\end{figure}

\subsection{Symmetry-Aware 6D Dataset (SA6D)}
\label{sec 3.1}
{\color{black} SA6D is designed to train WAPR with synthetic RGB-D pairs whose pose targets remain consistent under rotational symmetry. It starts from 944 GSO scans \cite{gso}, obtains KASAL-assisted rotational symmetry priors for these scans, and expands them through geometry and texture augmentation into \(\sim\)50K augmented object instances, including \(\sim\)30K rotationally symmetric instances. BlenderProc \cite{denninger_blenderproc_2019} is then used to render \(\sim\)2M synthetic RGB-D images.}

\paragraph{\textbf{Object Collection and Symmetry Augmentation}}\hspace{0pt}\\
{\color{black} The construction of SA6D starts from the original GSO scans. For each scan, human annotators specify the rotational symmetry type and order, and KASAL \cite{KASAL} localizes the corresponding symmetry-axis direction and center. The annotated scans are then expanded into augmented instances by geometry scaling and texture randomization, while reusing the corresponding symmetry priors. This avoids repeatedly annotating each augmented instance.}

{\color{black} SA6D uses two types of texture augmentation. The first modifies the texture while preserving the original texture-aware rotational symmetry prior, as illustrated by the disc in Fig. \ref{fig4}. The second replaces the texture with a uniform color, which removes texture-induced asymmetries and makes the augmented instance follow the geometry-only rotational symmetry prior, as shown by the carton in Fig. \ref{fig4}. Thus, each original scan is associated with both a texture-aware prior and a geometry-only prior. These priors are used in the offline canonicalization procedure described below to generate ambiguity-free candidate--target pose pairs for WAPR training.}

\begin{figure}[t]
	\centering
	\includegraphics[width=0.7\linewidth]{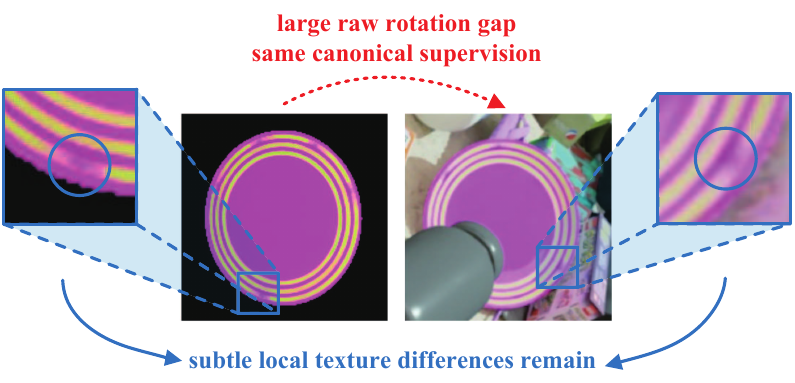}
	\caption{{\color{black} Rotational symmetry ambiguity in wide-angle training. The candidate and target views can have a large raw rotation gap, while their visible difference may be limited to subtle local texture details, as shown by the zoomed regions. With rotational symmetry priors, such pairs are treated as symmetry-equivalent and canonicalized to the same supervision before loss computation.
	}}
	\label{fig3}
\end{figure}

\paragraph{\textbf{Symmetry Disambiguation}}\hspace{0pt}\\
{\color{black} To train WAPR, candidate poses are simulated by applying random rotations and translations to ground-truth poses in SA6D. For rotationally symmetric objects, multiple symmetry-equivalent poses may be visually valid for the same training pair. Directly using the raw perturbed rotation can therefore assign inconsistent pose-update targets to visually similar candidate--target pairs, as illustrated in Fig. \ref{fig3}. SA6D resolves this ambiguity by applying an offline min-over-symmetry canonicalization before loss computation. The procedure selects a canonical representative from the symmetry-equivalent poses, making the pose update used for training unique and prediction-independent.}

Given a ground-truth pose, a set of augmented poses ${{\mathcal{P}}_{\text{aug}}}$ is generated by applying random perturbations to both rotation and translation:
\begin{equation}
	{{\mathcal{P}}_{\text{aug}}}=\{({{R}_{j}},{{t}_{j}})\mid j=1,2,\ldots ,M\}
\end{equation}

{\color{black} Each object is aligned so that its rotational symmetry center coincides with the origin, allowing symmetry transformations to be represented as pure rotations. For an object with \(n\)-fold discrete rotational symmetry, the symmetry group is represented by \(n\) rotation matrices. For continuous rotational symmetry, the symmetry orbit is discretized into 360 uniformly spaced rotations, which bounds the angular quantization error by \(0.5^\circ\). The symmetry-consistent rotation set is}

\begin{equation}
	\mathcal{R}_{\text{sym}} = 
	\{\, R_k \in \mathrm{SO}(3) \mid k = 1, 2, \dots, K \,\}
\end{equation}
where each $R_k$ represents a valid rotational symmetry of the object. 

For each augmented pose $({{R}_{j}},{{t}_{j}})$, symmetry-extended variants are constructed as:
\begin{equation}
	\mathcal{P}_{\text{sym}}^{(j)}=\{({{R}_{j}}{{R}_{k}},{{t}_{j}})\mid {{R}_{k}}\in {{\mathcal{R}}_{\text{sym}}}\}
\end{equation}

A canonical representative is selected by choosing the variant with the smallest rotation from the identity:
\begin{equation}
	(R_{\text{j}}^{*},t_{\text{j}}^{*})=\arg {{\min }_{(R,t)\in \mathcal{P}_{\text{sym}}^{(j)}}}\|R-I{{\|}_{2}}
\end{equation}

{\color{black} The selected pose \((R_{j}^{*},t_{j}^{*})\) is used as the final augmented candidate pose for training. The target pose is canonicalized under the same symmetry prior, yielding a consistent candidate--target pair with a unique rotation update for training. This canonicalization is performed when constructing SA6D training pairs and does not depend on the network prediction.}

\paragraph{\textbf{Segmentation Mask Generation}}\hspace{0pt}\\
{\color{black} WAPR also needs mask inputs during training. Instead of using perfect synthetic masks, which are cleaner than inference-time masks, SA6D generates training masks with a large-scale segmentation model such as SAM \cite{kirillov_sam_2023}. For each training sample, SAM produces candidate masks \(\mathcal{S}=\{S_i\}_{i=1}^{N}\). Masks whose intersection-over-union (IoU) with the ground-truth mask \(G\) exceeds 0.5 are kept:}
\begin{equation}
	\mathcal{Q}=\{i\mid \operatorname{IoU}({{S}_{i}},G)>0.5\}
\end{equation}

{\color{black} If \(\mathcal{Q}\) is non-empty, one mask is randomly sampled for training; otherwise, a blank mask is used. Together, symmetry-prior canonicalization and SAM-generated masks provide WAPR with well-defined pose targets and realistic mask inputs for wide-angle refinement training.}

\subsection{Wide-angle Pose Refinement}
\label{sec 3.2}
\paragraph{\textbf{Neural Network}}\hspace{0pt}\\

{\color{black} Given a candidate pose and its canonicalized target pose from SA6D, WAPR learns to predict the pose update by comparing two RGB-D observations. One branch takes the observed RGB-D crop and mask, which corresponds to a PBR-rendered sample during training and a real sensor observation during inference. The other branch takes the rendered RGB-D image and mask generated under the candidate pose. Two shared-weight ResNet layers based on BasicBlock-style residual units extract paired feature maps from the two branches.}

{\color{black} The paired feature maps are concatenated and augmented with positional encodings. A two-layer Transformer encoder models global interactions in the fused representation, as illustrated in Fig. \ref{fig2:twocol}. Two independent one-layer Transformer encoders then produce separate representations for rotation and translation. Each branch applies attention-based pooling and a linear layer to regress the corresponding pose update.}

{\color{black} Rotation updates are parameterized as 3D rotation vectors and converted to rotation matrices in \(\mathrm{SO}(3)\) before being applied to candidate poses.} For a candidate pose $({{R}_{i}},{{t}_{i}})$, the network predicts a rotation vector $\Delta {{\mathbf{\tilde{r}}}_{i}}\in {{\mathbb{R}}^{3}}$ and a translation offset $\Delta {{\tilde{t}}_{i}}\in {{\mathbb{R}}^{3}}$.

{\color{black} WAPR is trained and applied iteratively. During training, candidate poses are generated by perturbing the ground-truth poses with rotational deviations up to \(90^\circ\). After each iteration, the predicted rotation and translation updates are applied to the current candidate pose, and the updated pose is used to re-render the input for the next iteration.} The rotation and translation are updated by

\begin{equation}
	{{\tilde{R}}_{i}}=\operatorname{Rot}(\Delta {{\mathbf{\tilde{r}}}_{i}}){{R}_{i}}
\end{equation}
\begin{equation}
	{{\tilde{t}}_{i}}=\Delta {{\tilde{t}}_{i}}+{{t}_{i}}
\end{equation}
where $\operatorname{Rot}(\cdot )$ {\color{black} maps a 3D rotation vector to a rotation matrix in \(\mathrm{SO}(3)\).}

{\color{black} The \(90^\circ\) perturbation range is used as a practical wide-angle training range for render-and-compare refinement. As analyzed in the supplementary material, the Visual Correspondence Ratio (VCR) between the candidate and target views decreases with angular deviation, and very low-VCR pairs provide limited shared visual evidence for learning reliable updates. Perturbations far beyond \(90^\circ\), especially near \(180^\circ\), also make training convergence less stable. Therefore, \(90^\circ\) provides a wide yet learnable range for WAPR training.}

\paragraph{\textbf{Angle-Balanced Loss}}\hspace{0pt}\\
{\color{black} Wide-angle training contains both easy local refinements and hard large-error pairs. When the rendered and observed views have limited overlap, high-error samples can dominate the gradients while providing little useful correspondence. WAPR therefore uses an angle-balanced loss to emphasize learnable wide-angle cases and down-weight uninformative large-error samples. The prediction error of each sample is measured in rotation-vector space as the L1 residual magnitude between the predicted and target rotation-update vectors:}

\begin{equation}
	{{\theta}_{i}}=\left\|\Delta \tilde{\mathbf r}_{i}-\Delta \mathbf r_{i}\right\|_{1}
\end{equation}

Based on ${{\theta }_{i}}$, a dynamic weight ${{w}_{i}}$ is assigned to each training sample:
\begin{equation}
	{{w}_{i}}=\exp \left( \sigma \cdot \min \left\{ {{\theta }_{i}},\tau -{{\theta }_{i}} \right\} \right)
\end{equation}
{\color{black} where \(\tau\) is empirically set to \(0.25\pi\). Samples with rotation-vector residuals below \(\tau\) are treated as learnable wide-angle cases and receive larger weights, while samples with little overlap often yield \(\theta_i>\tau\) and are down-weighted. The coefficient \(\sigma\) controls the suppression strength and is set to 3.}

The final angle-balanced loss is then formulated as:
\begin{equation}
	L=\alpha \cdot \frac{B\cdot \sum\limits_{i=1}^{B}{{{w}_{i}}\cdot \left| \Delta \tilde{\mathbf r}_{i}-\Delta \mathbf r_{i} \right|}}{\sum\limits_{i=1}^{B}{{{w}_{i}}}}+\beta \cdot \sum\limits_{i=1}^{B}{\left| \Delta \tilde{\mathbf t}_{i}-\Delta \mathbf t_{i} \right|}
\end{equation}
where \(B\) denotes the number of samples in the mini-batch, \(\alpha\) and \(\beta\) are the balancing coefficients for the rotational and translational losses, respectively, and \(\left| \cdot \right|\) denotes the L1 loss. We set \(\alpha=3\) and \(\beta=1\).

\paragraph{\textbf{Inference Pipeline}}\hspace{0pt}\\
{\color{black} WAPR's wide-angle refinement ability allows the inference pipeline to use a sparse set of candidate poses. Given a 2D bounding box and mask, candidate poses are generated by combining a small set of viewing directions with evenly spaced in-plane rotations. In the fast setting, 12 candidates are formed by pairing the 4 viewing directions defined by the vertices of a regular tetrahedron with 3 in-plane rotations. In the unconstrained setting, 24 candidates are formed by pairing the 8 viewing directions defined by the vertices of a regular cube with the same 3 in-plane rotations.}

{\color{black} For each candidate, WAPR pairs the cropped RGB-D image and mask with rendered RGB-D and mask images generated under the candidate pose. In the unconstrained setting, all 24 candidates are refined for five rounds with re-rendered inputs at each round. In the fast setting, WAPR uses the two-stage schedule described in the supplementary material: all 12 candidates are first refined for two rounds and scored, after which only the top-4 candidates are refined for three additional rounds. Among the refined candidates, the pose with the highest confidence score predicted by the ScoreNet module of FoundationPose \cite{wen_foundationpose:_2024} is selected as the final result.}

%% file: sec/4_experiment.tex
\section{Experiments}
{\color{black} This section evaluates WAPR through ablation studies and comparisons with state-of-the-art methods \cite{Co-op,caraffa_freeze_2024,freezev2}. We first describe the experimental setup in Sec. \ref{sec 4.1}, then analyze the main design choices in Sec. \ref{sec 4.2}, and finally report BOP localization and detection results in Sec. \ref{sec 4.3}.}

\subsection{Experimental setup}
\label{sec 4.1}

\paragraph{\textbf{Network Settings}} \hspace{0pt}\\
{\color{black} Both input branches of WAPR use a fixed resolution of \(160\times160\). Each branch extracts features with a shared-weight encoder composed of a \(7\times7\) convolution, a \(3\times3\) convolution, and two ResNet basic blocks with 128 channels. The paired features are concatenated and processed by a fusion stage with 256- and 512-channel ResNet blocks, yielding a 512-dimensional fused feature representation. The Transformer encoder and attention-based pooling module both use four attention heads and a feed-forward dimension of 512. WAPR is trained for 800K iterations using Adam with a learning rate of 0.0002 and a batch size of 256. Each iteration samples RGB-D pairs from a single object category. Candidate poses are generated with translation offsets up to 50\% of the object diameter and rotational deviations up to \(90^\circ\), followed by five-step iterative refinement.}

\paragraph{\textbf{Datasets \& Metrics}} \hspace{0pt}\\
{\color{black} The experiments follow the BOP 2024 single-view model-based unseen-object protocol, where comparable baselines and default 2D detections are available. We evaluate on seven BOP-Classic-Core datasets: LM-O \cite{fleet_lmo_2014}, T-LESS \cite{hodan_t-less_2017}, ITODD \cite{drost_itodd_2017}, HB \cite{kaskman_hb_2019}, YCB-V \cite{xiang_posecnn_ycb-v_2018}, IC-BIN \cite{doumanoglou_ic-bin_2016}, and TUD-L \cite{ferrari_bop_tud-l_2018}. Following the BOP protocol \cite{hodan_bop_metric_2020,hodan_bop_2024}, we use the average of VSD, MSSD, and MSPD as the BOP score. We report average recall (AR) for 6D localization and average precision (AP) for 6D detection.}

\paragraph{\textbf{Fast inference}} \hspace{0pt}\\
For fast inference, WAPR uses the 12-candidate tetrahedron setting described in Sec.~\ref{sec 3.2}. For each MUSE detection, the associated SAM-based mask is used to estimate the initial translation by a depth-weighted average within the mask and the depth map. Detections are filtered using MUSE-predicted object labels and confidence scores with a threshold of 0.35. On an RTX 5090 system, this setting estimates poses for up to 25 detected object instances per second using the two-stage refinement and scoring schedule described in the supplementary material.

\paragraph{\textbf{Unconstrained inference}} \hspace{0pt}\\
When inference time is not constrained, WAPR uses the 24-candidate cube setting described in Sec.~\ref{sec 3.2}. When multiple 2D detectors are used, their detections are merged and NMS is applied to remove redundant boxes. For each retained detection, all category--detection--candidate combinations are evaluated without applying the confidence threshold used in the fast setting.

\begin{table}[!t]
	
	\centering
	\scriptsize
	\setlength{\tabcolsep}{12pt}
	\renewcommand{\arraystretch}{0.7}
	\caption{Analysis of candidate-pose coverage on IC-BIN. IC-BIN contains multiple identical objects in cluttered bin-like scenes, making it sensitive to the coverage of initial pose hypotheses. It is therefore used as a representative stress test for evaluating the number of candidate poses. ``Num'' denotes the number of candidate poses. Results are reported in AP for MSSD and MSPD, together with their mean.}
	\begin{tabular}{l|cccccc}
		\toprule
		Num   & 4     & 8     & 12    & 24    & 40    & 60 \\
		\midrule
		MSSD    & 59.6  & 67.3  & 71.4  & 71.9  & 72.8  & \textbf{73.3 } \\
		MSPD    & 54.2  & 63.1  & 68.9  & 69.8  & 70.4  & \textbf{70.9 } \\
		\midrule
		Mean    & 56.9  & 65.2  & 70.1  & 70.9  & 71.6  & \textbf{72.1 } \\
		\bottomrule
	\end{tabular}%
	
	\label{tab4}
\end{table}

\begin{table}[t]
	\centering
	\scriptsize
	\renewcommand{\arraystretch}{1.1}
	\caption{Ablations of WAPR components on BOP detection AP. Results are reported as AP for MSSD and MSPD on three representative datasets, together with their mean. IC-BIN, TUD-L, and LM-O are selected to cover crowded multi-instance scenes, large object motion and viewpoint variation, and heavy occlusion, respectively. ``w/o'' denotes ``without.'' ``WAPR (\(\pm90^\circ\))'' is trained with candidate poses obtained by randomly perturbing the ground-truth pose within \(\pm90^\circ\).}
	
	\begin{tabular}{l|l|cccccc|c}
		\toprule
		\multirow{2}[0]{*}{Row} & \multirow{2}[0]{*}{Method} & \multicolumn{2}{c}{IC-BIN \cite{doumanoglou_ic-bin_2016}} & \multicolumn{2}{c}{TUD-L \cite{ferrari_bop_tud-l_2018}} & \multicolumn{2}{c|}{LM-O \cite{fleet_lmo_2014}} & \multirow{2}[0]{*}{Mean} \\
		&    & MSSD & MSPD & MSSD & MSPD & MSSD & MSPD & \\
		\midrule
		A0  & WAPR(±90°) & 71.9 & 69.8 & 96.0 & 95.4 & 77.0 & 80.7 & 81.8 \\
		\midrule
		B0  & WAPR(±20°) & 61.5 & 58.0 & 77.4 & 76.5 & 68.0 & 71.0 & 68.7 \\
		B1  & WAPR(±40°) & 69.9 & 67.1 & 95.0 & 94.7 & 74.4 & 77.8 & 79.8 \\
		B2  & WAPR(±60°) & 71.7 & 68.9 & 95.6 & 95.1 & 74.9 & 78.3 & 80.8 \\
		B3  & WAPR(±140°) & 70.6 & 68.1 & 95.5 & 95.0 & 74.4 & 77.7 & 80.2 \\
		B4  & WAPR(±180°) & 63.3 & 60.2 & 89.6 & 89.2 & 72.9 & 76.2 & 75.2 \\
		\midrule
		C0  & A0 + w/o mask input & 71.0 & 68.5 & 95.2 & 94.9 & 75.6 & 78.9 & 80.7 \\
		C1  & A0 + train with GT mask & 61.8 & 59.4 & 94.3 & 93.9 & 73.9 & 77.2 & 76.8 \\
		\midrule
		D0  & A0 + w/o sym prior & 70.8 & 68.4 & 93.6 & 92.6 & 75.8 & 79.4 & 80.1 \\
		D1  & A0 + w/o angle-balanced Loss & 69.8 & 67.3 & 91.9 & 91.1 & 72.8 & 76.5 & 78.2 \\
		D2  & D1 + w/o sym prior & 61.8 & 58.0 & 91.0 & 91.0 & 69.3 & 73.0 & 74.0 \\
		D3    & A0 on GSO PBR (w/o SA6D) & 64.8  & 62.2  & 92.0   & 91.1  & 71.5  & 74.7  & 76.1 \\
		D4    & D0 on GSO PBR (w/o SA6D) & 63.9  & 61.1  & 90.6  & 89.7  & 69.3  & 72.6  & 74.5 \\
		\midrule
		E0  & A0 + w avg pool & 71.5 & 68.8 & 95.5 & 95.3 & 75.5 & 78.8 & 80.9 \\
		E1  & E0 + w/o transformer & 70.9 & 68.5 & 95.3 & 94.9 & 74.3 & 77.7 & 80.3 \\
		\bottomrule
	\end{tabular}
	
	\vspace{8pt}
	
	\begin{tabular}{l|l|rrrrrr|c|c}
		\toprule
		\multirow{2}[0]{*}{Row} & \multirow{2}[0]{*}{Method} & \multicolumn{2}{c}{IC-BIN} & \multicolumn{2}{c}{TUD-L} & \multicolumn{2}{c|}{LM-O} & \multirow{2}[0]{*}{Mean} & \multirow{2}[0]{*}{Time/s} \\
		&       & \multicolumn{1}{c}{MSSD} & \multicolumn{1}{c}{MSPD} & \multicolumn{1}{c}{MSSD} & \multicolumn{1}{c}{MSPD} & \multicolumn{1}{c}{MSSD} & \multicolumn{1}{c|}{MSPD} &       &  \\
		\midrule
		F0    & FreeZeV2 \cite{caraffa_freeze_2024, freezev2} & 73.4  & 70.6  & 98.6  & 99.1  & 78.5  & 81.9  & 83.7 & 27.8 \\
		F1    & F0 + C0 & 74.9  & 72.4  & 98.7  & 98.9  & 78.6  & 82.1  & 84.3 & 27.8 (+0.54) \\
		\midrule
		F2    & GigaPose\cite{nguyen2024gigaPose}+GenFlow\cite{moon_genflow_2024} & 42.0  & 38.4  & 71.4  & 71.7  & 56.1  & 59.5  & 56.5 & 4.5 \\
		F3    & F2+C0 & 43.4  & 40.0  & 72.7  & 73.0  & 57.3  & 60.2  & 57.8 & 4.5 (+0.37) \\
		\bottomrule
	\end{tabular}
	
	\label{tab2}
\end{table}

\subsection{Ablation Study on Pose Refinement}
\label{sec 4.2}
{\color{black} This section analyzes the main factors affecting wide-angle pose refinement, including candidate-pose coverage, training inputs, symmetry priors, SA6D, angle-balanced optimization, network architecture, and plug-in refinement. The ablations in Table \ref{tab2} are conducted on three representative datasets for controlled component analysis, while the final all-seven-dataset evaluation is reported in Table \ref{tab5}.}

\paragraph{\textbf{Number of Candidate Poses}}\hspace{0pt}\\
{\color{black} Table \ref{tab4} studies how candidate-pose coverage affects WAPR on IC-BIN. Increasing the number of candidates improves the chance that at least one hypothesis falls within the effective refinement range. The gain is large when increasing from 4 to 12 candidates, where the mean AP improves from 56.9\% to 70.1\%. Further increasing the number to 24, 40, or 60 provides smaller gains, indicating that 12 candidates already provide a strong accuracy--efficiency trade-off for the fast setting, while 24 candidates are used in the unconstrained setting for higher coverage.}

\paragraph{\textbf{Angle Range for Training}}\hspace{0pt}\\
{\color{black} Rows B0--B4 in Table \ref{tab2} evaluate the rotation range used to generate candidate poses during training. Small perturbation ranges such as \(\pm20^\circ\) provide insufficient coverage for wide-angle errors, while overly large ranges such as \(\pm180^\circ\) introduce many low-correspondence samples and make optimization less stable. The \(\pm90^\circ\) setting achieves the best overall performance, consistent with the VCR and training-convergence analysis discussed in Sec. \ref{sec 3.2}.}

\paragraph{\textbf{Mask Input}}
{\color{black} Rows C0 and C1 evaluate the role of mask inputs. Removing the mask slightly degrades performance, showing that masks provide useful object-region guidance during render-and-compare refinement. Training with perfect ground-truth masks leads to a larger drop, since such masks differ from the predicted masks used at inference time. SAM-generated masks therefore provide a more realistic training signal for WAPR.}

\paragraph{\textbf{Rotational Symmetry Priors and SA6D Dataset}}
{\color{black} Rows D0--D4 evaluate symmetry priors and SA6D through controlled comparisons. Rows D1 and D2 isolate the effect of rotational symmetry priors without angle-balanced loss, where adding the priors improves AP by 4.2\%. Rows A0 and D0 show the effect when angle-balanced loss is used, where the priors improve AP by 1.7\%. Rows D3 and D4 train the corresponding configurations on original GSO PBR data without SA6D augmentation. Compared with these GSO-PBR counterparts, A0 and D0 improve AP by 5.7\% and 5.6\%, respectively, showing the benefit of SA6D training data under matched model configurations.}

\paragraph{\textbf{Angle-Balanced Loss Function}}
{\color{black} Rows D0--D2 analyze angle-balanced optimization. Compared with D2, which uses neither angle-balanced loss nor symmetry priors, D0 improves AP by 6.1\%, showing that angle-balanced loss stabilizes wide-angle training. Combining angle-balanced loss with symmetry priors further improves performance, with A0 outperforming D2 by 7.8\%.}

\paragraph{\textbf{Network Architecture}}
{\color{black} Rows E0 and E1 ablate the feature aggregation modules. Replacing attention-based pooling with average pooling reduces AP by 0.9\%, and further removing the Transformer encoder causes an additional 0.6\% drop, resulting in a total 1.5\% drop from A0. These results indicate that global feature interaction and attention-based aggregation are both useful for wide-angle render-and-compare refinement.}

\paragraph{\textbf{Plug-in Refinement}}
{\color{black} Rows F0--F3 evaluate the mask-free WAPR variant C0 as a plug-in refiner for existing pose estimators. When applied to FreeZeV2 poses, C0 improves AP by 0.6\% with a 0.54 s per-image overhead on top of FreeZeV2's 27.8 s runtime. When applied to GigaPose+GenFlow poses, C0 improves AP by 1.3\% with 0.37 s additional time. These results show that WAPR-style wide-angle refinement can further improve strong existing estimators, although the remaining gain is naturally smaller when the initial estimates are already strong.}

\paragraph{\textbf{Comparison with Refinement Baselines}}
{\color{black} Table \ref{tab4-1} compares WAPR with representative RGB-D refinement baselines on LM-O and YCB-V. This comparison complements the full-pipeline BOP evaluation in Table \ref{tab5} by focusing on refinement behavior under comparable settings. Compared with OSOP+ICP, WAPR is 4\(\times\) faster and improves mean AR by 29.7\%; compared with PPF/SIFT+Zephyr, it improves AR by 26.7\%. Under the same hypothesis setting, replacing WAPR with ICP reduces AR by 54.2\%, reflecting the sensitivity of ICP to large hypothesis errors. Compared with FoundationPose, which uses small-angle refinement with a maximum correction of \(22.5^\circ\), WAPR improves AR by 8.3\%.}

\begingroup
\setlength{\intextsep}{6pt}
\setlength{\textfloatsep}{6pt}
\begin{table}[H]
	
	\centering
	\scriptsize
	\setlength{\tabcolsep}{6pt}
	\setlength{\abovecaptionskip}{0pt}
	\setlength{\belowcaptionskip}{1pt}
	\renewcommand{\arraystretch}{1.03}
	\caption{Comparison with representative RGB-D refinement baselines on LM-O and YCB-V, where comparable prior-refinement results are available. The last three rows use MUSE detections and 12 hypotheses per instance.}
	\begin{tabular*}{\linewidth}{@{\extracolsep{\fill}}l|cc|c|c}
		\toprule
		Method & LM-O  & YCB-V & Mean  & Time (s/img) \\
		\midrule
		OSOP \cite{OSOP} + ICP & 48.2  & 57.2  & 52.7  & 4 \\
		(PPF \cite{PPF}, SIFT) + Zephyr \cite{Zephyr} & 59.8  & 51.6  & 55.7  & --  \\
		MegaPose-RGBD \cite{labbe_megapose_2022} & 58.3  & 63.3  & 60.8  & --  \\
		\midrule
		FoundationPose \cite{wen_foundationpose:_2024} (MUSE) + 12 hypotheses & 67.4  & 80.8  & 74.1  & --  \\
		ICP (MUSE) + 12 hypotheses & 26.6  & 29.7  & 28.2 & 35 \\
		WAPR (MUSE) + 12 hypotheses & \textbf{75.5}  & \textbf{89.3}  & \textbf{82.4}  & \textbf{1} \\
		\bottomrule
	\end{tabular*}
	
	\label{tab4-1}
\end{table}
\endgroup

\begingroup
\setlength{\intextsep}{6pt}
\setlength{\textfloatsep}{6pt}
\begin{table}[H]
	\centering
	\scriptsize
	\setlength{\abovecaptionskip}{12pt}
	\setlength{\belowcaptionskip}{1pt}
	\renewcommand{\arraystretch}{1.03}
	\caption{Comparison with SOTA methods on seven BOP-Classic-Core datasets for 6D localization (AR, top) and detection (AP, bottom). ``2D Det.'' reports the detector or hypothesis source. WAPR is trained on SA6D, while other methods use original training or released settings. ``Multi-Det.'' combines NIDS \cite{nids}, MUSE \cite{muse}, SAM6D \cite{lin_sam-6d:_2024}, and CNOS \cite{nguyen_cnos_2023}. FP denotes FoundationPose \cite{wen_foundationpose:_2024}. ``Time'' reports per-image runtime, and ``*'' marks proposed methods.}
	\begin{tabular*}{\linewidth}{@{\extracolsep{\fill}}c|l|l|ccccccc|c|c}
		\toprule
		\multicolumn{12}{c}{Model-based 6D localization of unseen objects – BOP-Classic-Core} \\
		\midrule
		& Method & 2D Det. & LM-O  & T-LESS & TUD-L & IC-BIN & ITODD & HB    & YCB-V & Mean  & Time (s) \\
		\midrule
		\multirow{2}[0]{*}{\rotatebox{90}{\scalebox{0.85}[1]{\textit{fast}}}} & Co-op  \cite{Co-op} & F3DT2D & 73.0    & 68.0    & 92.9  & 62.4  & 60.0    & 86.3  & 88.6  & 75.9  & 0.765 \\
		& WAPR* & F3DT2D & \textbf{75.4}  & 72.0  & 92.5  & 66.3  & 65.0  & 86.7  & \textbf{89.1}  & 78.1  & 0.904 \\
		& WAPR* & MUSE  & \textbf{75.4}  & \textbf{73.1}  & \textbf{96.1}  & \textbf{70.9}  & \textbf{70.5}  & \textbf{90.1}  & 88.3  & \textbf{80.6}  & 0.72 \\
		\midrule
		\multirow{8}[0]{*}{\rotatebox{90}{\scalebox{0.68}[1]{\textit{unconstrained}}}} & MegaPose \cite{labbe_megapose_2022} & CNOS  & 62.6 & 48.7 & 85.1 & 46.7 & 46.8 & 73.0 & 76.4 & 62.8 & 141.965 \\
		& Genflow \cite{moon_genflow_2024} & CNOS  & 67.8 & 55.6 & 81.1 & 56.3 & 57.5 & 79.1 & 82.5 & 68.6 & 11.14 \\
		& SAM6D \cite{lin_sam-6d:_2024} & SAM6D & 69.9 & 51.5 & 90.4 & 58.8 & 60.2 & 77.6 & 84.5 & 70.4 & 4.367 \\
		& FP \cite{wen_foundationpose:_2024} & SAM6D & 75.6 & 64.6 & 92.3 & 50.8 & 58.0 & 83.5 & 88.9 & 73.4 & 29.317 \\
		& Co-op  \cite{Co-op} & F3DT2D & 73.8 & 69.5 & \textbf{92.9} & 63.5 & 62.9 & 87.8 & 89.8 & 77.1 & 6.924 \\
		& WAPR* & CNOS  & \textbf{76.8} & \textbf{82.6} & 92.4 & \textbf{70.7} & \textbf{74.3} & \textbf{89.1} & \textbf{91.1} & \textbf{82.4} & 43.068 \\
		\cmidrule(l{1.0em}r){1-12}
		& FreeZeV2 \cite{caraffa_freeze_2024, freezev2} & Multi-Det. & 77.7 & 79.9 & \textbf{97.5} & 71.1 & 75.4 & 89.5 & \textbf{91.8} & 83.3 & 21.709 \\
		& WAPR* & Multi-Det. & \textbf{78.1}  & \textbf{81.7}  & 96.1  & \textbf{73.5}  & \textbf{79.8}  & \textbf{91}    & 91.5  & \textbf{84.5}  & 50.816 \\
		\midrule
		\multicolumn{12}{c}{Model-based 6D detection of unseen objects – BOP-Classic-Core} \\
		\midrule
		& Method & 2D Det. & LM-O  & T-LESS & TUD-L & IC-BIN & ITODD & HB    & YCB-V & Mean  & Time (s) \\
		\midrule
		\multirow{2}[0]{*}{\rotatebox{90}{\scalebox{0.85}[1]{\textit{fast}}}} & Co-op  \cite{Co-op} & F3DT2D  & 69.8 & 62.0 & 84.1 & 56.4 & 57.6 & 74.6 & 80.8 & 69.3 & 0.865 \\
		& WAPR* & F3DT2D & 74.7 & \textbf{64.5} & 96.3 & 67.2 & 68.1 & 77.6 & 87.1 & 76.5 & 0.987 \\
		& WAPR* & MUSE  & \textbf{75.6}  & 63.6  & \textbf{98.6}  & \textbf{71.8}  & \textbf{71.5}  & \textbf{81.6}  & \textbf{87.6}  & \textbf{78.6}  & 0.828 \\
		\midrule
		\multirow{5}[0]{*}{\rotatebox{90}{\scalebox{0.56}[1]{\textit{unconstrained}}}} & Genflow \cite{moon_genflow_2024} & CNOS  & 59.7 & 56.5 & 68.8 & 43.9 & 42.5 & 70.7 & 60.7 & 57.5 & 15.486 \\
		& Co-op \cite{Co-op} & CNOS  & 68.3  & 59.6  & 80.8  & 46.9  & 56.0    & 74.3  & 78.2  & 66.3  & 2.202 \\
		& WAPR* & CNOS  & \textbf{78.8} & \textbf{82.9} & \textbf{95.7} & \textbf{70.9} & \textbf{77.6} & \textbf{86.3} & \textbf{91.1} & \textbf{83.3} & 43.068 \\
		\cmidrule(l{1.0em}r){1-12}
		& FreeZeV2 \cite{caraffa_freeze_2024, freezev2} & Multi-Det. & 80.2 & 77.7 & \textbf{98.8} & 72.0 & 77.5 & 85.4 & \textbf{91.1} & 83.3 & 27.776 \\
		& WAPR* &  Multi-Det. & \textbf{80.7}  & \textbf{82.5}  & 98.7  & \textbf{75.0}    & \textbf{82.2}  & \textbf{88.7}  & 89.6  & \textbf{85.3}  & 50.816 \\
		\bottomrule
	\end{tabular*}
	\label{tab5}
\end{table}
\endgroup

\FloatBarrier

\subsection{BOP Results on 6D Localization and Detection}
\label{sec 4.3}

{\color{black} Table \ref{tab5} compares WAPR with state-of-the-art methods on seven BOP-Classic-Core datasets for 6D localization and detection. In the unconstrained Multi-Det. setting, WAPR achieves 84.5\% AR and 85.3\% AP, improving over FreeZeV2 by 1.2\% AR and 2.0\% AP. This accuracy-oriented setting uses a larger candidate budget and obtains higher mean accuracy with a longer per-image runtime. The fast setting reports the efficiency-oriented operating point under the \(<1\) s per-image budget.}

{\color{black} In the fast setting, WAPR with MUSE detections achieves 80.6\% AR and 78.6\% AP, improving over Co-op by 4.7\% AR and 9.3\% AP. For a matched-detector comparison, WAPR is also evaluated with the same F3DT2D detector used by Co-op under the same fast-inference protocol. This variant achieves 78.1\% AR and 76.5\% AP, improving over Co-op by 2.2\% AR and 7.2\% AP. These results show that the gain is preserved under the same detector, while the MUSE-based variant gives the highest mean performance among the evaluated fast-setting variants.}

%% file: sec/5_conclusion.tex
\section{Conclusion}
\enlargethispage{6\baselineskip}
{\color{black} This paper presented WAPR, a zero-shot wide-angle pose refinement model for unseen-object 6D pose estimation. WAPR refines sparse candidate poses by comparing real and rendered RGB-D observations with masks, enabling accurate pose estimation under large candidate-pose deviations. To support wide-angle training, this paper further constructed SA6D, a large-scale synthetic RGB-D dataset with KASAL-assisted rotational symmetry priors. These priors are used to canonicalize symmetry-equivalent candidate--target pose pairs, while angle-balanced loss stabilizes learning across different angular ranges. Experiments on seven BOP core datasets show that WAPR achieves strong performance in both fast and unconstrained settings. These results demonstrate that wide-angle refinement, scalable training data with symmetry priors, and sparse candidate evaluation provide an effective solution for unseen-object 6D pose estimation.}

%% file: sec/Acknowledgments.tex
\section*{Acknowledgments}
This work was supported by the National Natural Science Foundation of China under Grant No.~52375487, the Special Fund of Jiangsu Province for Key Research and Development under Grant No.~BE2023041, and the Doctoral Research Innovation Ability Improvement Program of Southeast University under Grant No.~CXJH\_SEU~26032. The authors sincerely thank the area chairs and anonymous reviewers for their careful reading, constructive feedback, and detailed suggestions.

%% file: main.bib
@String(PAMI = {IEEE Trans. Pattern Anal. Mach. Intell.})

@String(IJCV = {Int. J. Comput. Vis.})

@String(CVPR= {IEEE Conf. Comput. Vis. Pattern Recog.})

@String(ICCV= {Int. Conf. Comput. Vis.})

@String(ECCV= {Eur. Conf. Comput. Vis.})

@String(TIP  = {IEEE Trans. Image Process.})

@String(CVPRW= {IEEE Conf. Comput. Vis. Pattern Recog. Worksh.})

@String(VR   = {Vis. Res.})

@String(PAMI  = {IEEE TPAMI})

@String(IJCV  = {IJCV})

@String(CVPR  = {CVPR})

@String(ICCV  = {ICCV})

@String(ECCV  = {ECCV})

@String(TIP   = {IEEE TIP})

@String(CVPRW= {CVPRW})

@String(IROS = {IROS})

@String(VISIGRAPP = {VISIGRAPP})

@article{liu_gdrnpp_2022,
	author = {Liu, Xingyu and Zhang, Ruida and Zhang, Chenyangguang and Wang, Gu and Tang, Jiwen and Li, Zhigang and Ji, Xiangyang},
	title = {{GDRNPP}: A Geometry-Guided and Fully Learning-Based Object Pose Estimator},
	journal = {PAMI},
	volume = {47},
	number = {7},
	year = {2025},
	pages = {5742 - 5759},
}

@inproceedings{Co-op,
	author = {Moon, Sungphill and Son, Hyeontae and Hur, Dongcheol and Kim, Sangwook},
	title = {Co-op: Correspondence-based Novel Object Pose Estimation},
	booktitle = {CVPR},
	pages = {11622 - 11632},
	year = 2025
}

@inproceedings{OSOP,
	author = {Shugurov, Ivan and Li, Fu and Busam, Benjamin and Ilic, Slobodan},
	title = {OSOP: A Multi-Stage One Shot Object Pose Estimation Framework},
	booktitle = {CVPR},
	pages = {6825-6834},
	year = 2022
}

@inproceedings{PPF,
	author = {Drost, Bertram and Ulrich, Markus and Navab, Nassir and Ilic, Slobodan},
	title = {Model globally, match locally: Efficient and robust 3D object recognition},
	booktitle = {CVPR},
	pages = {998-1005},
	year = 2010
}

@inproceedings{Zephyr,
	author = {Brian Okorn and Qiao Gu and Martial Hebert and David Held},
	title = {ZePHyR: Zero-shot Pose Hypothesis Rating},
	booktitle = {ICRA},
	pages = {14141-14148},
	year = 2021
}

@misc{nids,
	title={Adapting Pre-Trained Vision Models for Novel Instance Detection and Segmentation}, 
	author={Yangxiao Lu and Jishnu Jaykumar P and Yunhui Guo and Nicholas Ruozzi and Yu Xiang},
	year=2024,
	note = {ArXiv preprint arXiv:2405.17859},
}

@InProceedings{HccePose_BF,
	author    = {Wang, Yulin and Hu, Mengting and Li, Hongli and Luo, Chen},
	title     = {{HccePose(BF)}: Predicting Front \& Back Surfaces to Construct Ultra-Dense {2D}-{3D} Correspondences for Pose Estimation},
	booktitle = {ICCV},
	year      = {2025},
	pages     = {7166-7175}
}

@ARTICLE{KASAL,
	author = {Wang, Yulin and Luo, Chen},
	title  = {Key-Axis-Based Localization of Symmetry Axes in 3D Objects Utilizing Geometry and Texture}, 
	journal= {TIP}, 
	year   = {2024},
	volume = {33},
	pages  = {6720-6733},
}

@misc{denninger_blenderproc_2019,
	author = {Denninger, Maximilian and Sundermeyer, Martin and Winkelbauer, Dominik and Zidan, Youssef and Olefir, Dmitry and Elbadrawy, Mohamad and Lodhi, Ahsan and Katam, Harinandan},
	title = {BlenderProc},
	note = {ArXiv preprint arXiv:1911.01911},
	year = 2019
}

@misc{ge_yolox_2021,
	author = {Ge, Zheng and Liu, Songtao and Wang, Feng and Li, Zeming and Sun, Jian},
	title = {{YOLOX}: Exceeding {YOLO} Series in 2021},
	note = {ArXiv preprint arXiv:2107.08430},
	year = 2021
}

@misc{yu_robotic2_2024,
	author = {Yu, Sheng and Zhai, Di-Hua and Guan, Yuyin and Xia, Yuanqing},
	title = {Category-Level 6-D Object Pose Estimation With Shape Deformation for Robotic Grasp Detection},
	note = {Early access in \emph{IEEE TNNLS}.},
	year = 2024
}

@article{ren_fast-rcnn_2017,
	author = {Ren, Shaoqing and He, Kaiming and Girshick, Ross and Sun, Jian},
	title = {Faster R-CNN: Towards Real-Time Object Detection with Region Proposal Networks},
	journal = {PAMI},
	volume = 39,
	number = 6,
	pages = {1137--1149},
	year = 2017
}

@article{yang_ar3_2023,
	author = {Yang, Xu and Cai, Junqi and Li, Kunbo and Fan, Xiumin and Cao, Hengling},
	title = {A monocular-based tracking framework for industrial augmented reality applications},
	journal = {The International Journal of Advanced Manufacturing Technology},
	volume = 128,
	number = 5,
	pages = {2571--2588},
	year = 2023
}

@article{hoque_ad1_2023,
	author = {Hoque, Sabera and Xu, Shuxiang and Maiti, Ananda and Wei, Yuchen and Arafat, Md. Yasir},
	title = {Deep Learning for {6D} Pose Estimation of Objects — A Case Study for Autonomous Driving},
	journal = {Expert Syst. Appl.},
	volume = 223,
	number = 119838,
	pages = {1--15},
	year = 2024
}

@article{park_ar1_2024,
	author = {Park, Kyeong-Beom and Choi, Sung Ho and Lee, Jae Yeol},
	title = {Self-training based augmented reality for robust 3D object registration and task assistance},
	journal = {Expert Syst. Appl.},
	volume = 238,
	number = 122331,
	pages = {1--16},
	year = 2024
}

@article{zou_ad2_2021,
	author = {Zou, Wenbin and Wu, Di and Tian, Shishun and Xiang, Canqun and Li, Xia and Zhang, Lu},
	title = {End-to-End {6DoF} Pose Estimation From Monocular {RGB} Images},
	journal = {IEEE TCE},
	volume = 67,
	number = 1,
	pages = {87--96},
	year = 2021
}

@article{liu_robotic1_2023,
	author = {Liu, Jian and Sun, Wei and Liu, Chongpei and Zhang, Xing and Fu, Qiang},
	title = {Robotic Continuous Grasping System by Shape Transformer-Guided Multiobject Category-Level 6-D Pose Estimation},
	journal = {IEEE TII},
	volume = 19,
	number = 11,
	pages = {11171--11181},
	year = 2023
}

@article{zhang_robotic3_2022,
	author = {Zhang, Hui and Liang, Zhicong and Li, Chen and Zhong, Hang and Liu, Li and Zhao, Chenyang and Wang, Yaonan and Wu, Q. M. Jonathan},
	title = {A Practical Robotic Grasping Method by Using 6-D Pose Estimation With Protective Correction},
	journal = {IEEE TIE},
	volume = 69,
	number = 14,
	pages = {3876--3886},
	year = 2022
}

@article{thalhammer_robotic4_2024,
	author = {Thalhammer, Stefan and Bauer, Dominik and Hönig, Peter and Weibel, Jean-Baptiste and García-Rodríguez, José and Vincze, Markus},
	title = {Challenges for Monocular 6-D Object Pose Estimation in Robotics},
	journal = {IEEE TRO},
	volume = 40,
	number = 1,
	pages = {4065--4084},
	year = 2024
}

@article{peng_pvnet:_2018,
	author = {Peng, Sida and Zhou, Xiaowei and Liu, Yuan and Lin, Haotong and Huang, Qixing and Bao, Hujun},
	title = {{PVNet}: Pixel-Wise Voting Network for {6DoF} Object Pose Estimation},
	journal = {PAMI},
	volume = 44,
	number = 6,
	pages = {3212--3223},
	year = 2022
}

@inproceedings{nguyen_bop_2025,
	title={{BOP} Challenge 2024 on Model-Based and Model-Free {6D} Object Pose Estimation},
	author={Nguyen, Van Nguyen and Tyree, Stephen and Guo, Andrew and Fourmy, Mederic and Gouda, Anas and Lee, Taeyeop and Moon, Sungphill and Son, Hyeontae and Ranftl, Lukas and Tremblay, Jonathan and Brachmann, Eric and Drost, Bertram and Lepetit, Vincent and Rother, Carsten and Birchfield, Stan and Matas, Jiri and Labbe, Yann and Sundermeyer, Martin and Hodan, Tomas},
	booktitle = {CVPRW},
	year={2025}
}

@inproceedings{hodan_bop_2024,
	author = {Hodan, Tomas and Sundermeyer, Martin and Labbe, Yann and Nguyen, Van Nguyen and Wang, Gu and Brachmann, Eric and Drost, Bertram and Lepetit, Vincent and Rother, Carsten and Matas, Jiri},
	title = {{BOP} Challenge 2023 on Detection, Segmentation and Pose Estimation of Seen and Unseen Rigid Objects},
	booktitle = {CVPRW},
	pages = {5610--5619},
	year = 2024
}

@inproceedings{lin_sam-6d:_2024,
	author = {Lin, Jiehong and Liu, Lihua and Lu, Dekun and Jia, Kui},
	title = {{SAM-6D}: Segment Anything Model Meets Zero-Shot {6D} Object Pose Estimation},
	booktitle = {CVPR},
	pages = {27906--27916},
	year = 2024
}

@inproceedings{wen_foundationpose:_2024,
	author = {Wen, Bowen and Yang, Wei and Kautz, Jan and Birchfield, Stan},
	title = {{FoundationPose}: Unified {6D} Pose Estimation and Tracking of Novel Objects},
	booktitle = {CVPR},
	pages = {17868--17879},
	year = 2024
}

@inproceedings{hodan_epos:_2020,
	author = {Hodan, Tomas and Barath, Daniel and Matas, Jiri},
	title = {{EPOS}: Estimating {6D} Pose of Objects With Symmetries},
	booktitle = {CVPR},
	pages = {11700--11709},
	year = 2020
}

@inproceedings{wang_gdr-net:_2021,
	author = {Wang, Gu and Manhardt, Fabian and Tombari, Federico and Ji, Xiangyang},
	title = {{GDR-Net}: Geometry-Guided Direct Regression Network for Monocular {6D} Object Pose Estimation},
	booktitle = {CVPR},
	pages = {16606--16616},
	year = 2021
}

@inproceedings{su_zebrapose:_2022,
	author = {Su, Yongzhi and Saleh, Mahdi and Fetzer, Torben and Rambach, Jason and Navab, Nassir and Busam, Benjamin and Stricker, Didier and Tombari, Federico},
	title = {{ZebraPose}: Coarse to Fine Surface Encoding for {6DoF} Object Pose Estimation},
	booktitle = {CVPR},
	pages = {6728--6738},
	year = 2022
}

@inproceedings{li_deepim:_2020,
	author = {Li, Yi and Wang, Gu and Ji, Xiangyang and Xiang, Yu and Fox, Dieter},
	title = {{DeepIM}: Deep Iterative Matching for {6D} Pose Estimation},
	booktitle = {ECCV},
	pages = {695--711},
	year = 2018
}

@inproceedings{zakharov_dpod:_2019,
	author = {Zakharov, Sergey and Shugurov, Ivan and Ilic, Slobodan},
	title = {{DPOD}: {6D} Pose Object Detector and Refiner},
	booktitle = {ICCV},
	pages = {1941--1950},
	year = 2019
}

@inproceedings{kirillov_sam_2023,
	author = {Kirillov, Alexander and Mintun, Eric and Ravi, Nikhila and Mao, Hanzi and Rolland, Chloe and Gustafson, Laura and Xiao, Tete and Whitehead, Spencer and Berg, Alexander C. and Lo, Wan-Yen and Dollár, Piotr and Girshick, Ross},
	title = {Segment Anything},
	booktitle = {ICCV},
	pages = {3992--4003},
	year = 2023
}

@inproceedings{nguyen_cnos_2023,
	author = {Nguyen, Van Nguyen and Groueix, Thibault and Ponimatkin, Georgy and Lepetit, Vincent and Hodan, Tomas},
	title = {{CNOS}: A Strong Baseline for {CAD}-based Novel Object Segmentation},
	booktitle = {ICCVW},
	pages = {2126--2132},
	year = 2023
}

@misc{muse,
	author = {Sungmin Cho and Sungbum Park and Insoo Oh},
	title = {MUSE: Model-based Uncertainty-aware Similarity Estimation for zero-shot 2D Object Detection and Segmentation},
	note = {ArXiv preprint arXiv:2510.17866},
	year = 2025
}

@inproceedings{tian_fcos_2019,
	author = {Tian, Zhi and Shen, Chunhua and Chen, Hao and He, Tong},
	title = {FCOS: Fully Convolutional One-Stage Object Detection},
	booktitle = {ICCV},
	pages = {9626--9635},
	year = 2019
}

@inproceedings{labbe_megapose_2022,
	author = {Labbé, Yann and Manuelli, Lucas and Mousavian, Arsalan and Tyree, Stephen and Birchfield, Stan and Tremblay, Jonathan and Carpentier, Justin and Aubry, Mathieu and Fox, Dieter and Sivic, Josef},
	title = {{MegaPose}: {6D} Pose Estimation of Novel Objects via Render \& Compare},
	booktitle = {CoRL},
	pages = {1--10},
	year = 2022
}

@inproceedings{xiang_posecnn_ycb-v_2018,
	author = {Xiang, Yu and Schmidt, Tanner and Narayanan, Venkatraman and Fox, Dieter},
	title = {{PoseCNN}: A Convolutional Neural Network for {6D} Object Pose Estimation in Cluttered Scenes},
	booktitle = {RSS},
	pages = {1--10},
	year = 2018
}

@inproceedings{he_resnet_2015,
	author = {He, Kaiming and Zhang, Xiangyu and Ren, Shaoqing and Sun, Jian},
	title = {Deep Residual Learning for Image Recognition},
	booktitle = {CVPR},
	pages = {770--778},
	year = 2016
}

@inproceedings{fleet_lmo_2014,
	author = {Brachmann, Eric and Krull, Alexander and Michel, Frank and Gumhold, Stefan and Shotton, Jamie and Rother, Carsten},
	title = {Learning {6D} Object Pose Estimation Using {3D} Object Coordinates},
	booktitle = {ECCV},
	pages = {536--551},
	year = 2014
}

@inproceedings{hodan_t-less_2017,
	author = {Hodan, Tomas and Haluza, Pavel and Obdrzalek, Stepan and Matas, Jiri and Lourakis, Manolis and Zabulis, Xenophon},
	title = {{T-LESS}: An {RGB-D} Dataset for {6D} Pose Estimation of Texture-Less Objects},
	booktitle = {WACV},
	pages = {880--888},
	year = 2017
}

@inproceedings{drost_itodd_2017,
	author = {Drost, Bertram and Ulrich, Markus and Bergmann, Paul and Hartinger, Philipp and Steger, Carsten},
	title = {Introducing {MVTec} {ITODD} — A Dataset for {3D} Object Recognition in Industry},
	booktitle = {ICCVW},
	pages = {2200--2208},
	year = 2017
}

@inproceedings{kaskman_hb_2019,
	author = {Kaskman, Roman and Zakharov, Sergey and Shugurov, Ivan and Ilic, Slobodan},
	title = {{HomebrewedDB}: {RGB-D} Dataset for {6D} Pose Estimation of {3D} Objects},
	booktitle = {ICCVW},
	pages = {2767--2776},
	year = 2019
}

@inproceedings{doumanoglou_ic-bin_2016,
	author = {Doumanoglou, Andreas and Kouskouridas, Rigas and Malassiotis, Sotiris and Kim, Tae-Kyun},
	title = {Recovering {6D} Object Pose and Predicting Next-Best-View in the Crowd},
	booktitle = {CVPR},
	pages = {3583--3592},
	year = 2016
}

@inproceedings{ferrari_bop_tud-l_2018,
	author = {Hodan, Tomas and Michel, Frank and Brachmann, Eric and Kehl, Wadim and Buch, Anders Glent and Kraft, Dirk and Drost, Bertram and Vidal, Joel and Ihrke, Stephan and Zabulis, Xenophon and Sahin, Caner and Manhardt, Fabian and Tombari, Federico and Kim, Tae-Kyun and Matas, Jiri and Rother, Carsten},
	title = {{BOP}: Benchmark for {6D} Object Pose Estimation},
	booktitle = {ECCV},
	pages = {19--35},
	year = 2018
}

@inproceedings{hodan_bop_metric_2020,
	author = {Hodan, Tomas and Sundermeyer, Martin and Drost, Bertram and Labbe, Yann and Brachmann, Eric and Michel, Frank and Rother, Carsten and Matas, Jiri},
	title = {{BOP} Challenge 2020 on {6D} Object Localization},
	booktitle = {ECCV},
	pages = {577--594},
	year = 2020
}

@inproceedings{he_maskrcnn_2017,
	author = {He, Kaiming and Gkioxari, Georgia and Dollar, Piotr and Girshick, Ross},
	title = {Mask R-CNN},
	booktitle = {ICCV},
	pages = {2980--2988},
	year = 2017
}

@inproceedings{li_ar2_2024,
	author = {Li, Shiyu and Schieber, Hannah and Corell, Niklas and Egger, Bernhard and Kreimeier, Julian and Roth, Daniel},
	title = {GBOT: Graph-Based 3D Object Tracking for Augmented Reality-Assisted Assembly Guidance},
	booktitle = {VR},
	pages = {513--523},
	year = 2024
}

@inproceedings{wu_ad3_2019,
	author = {Wu, Di and Zhuang, Zhaoyong and Xiang, Canqun and Zou, Wenbin and Li, Xia},
	title = {{6D-VNet}: End-To-End {6DoF} Vehicle Pose Estimation From Monocular {RGB} Images},
	booktitle = {CVPRW},
	pages = {1238--1247},
	year = 2019
}

@inproceedings{hong_rdpn6d_2024,
	author = {Hong, Zong-Wei and Hung, Yen-Yang and Chen, Chu-Song},
	title = {{RDPN6D}: Residual-based Dense Point-wise Network for {6DoF} Object Pose Estimation Based on {RGB-D} Images},
	booktitle = {CVPRW},
	pages = {5251--5260},
	year = 2024
}

@inproceedings{hai_radet_2023,
	author = {Hai, Yang and Song, Rui and Li, Jiaojiao and Salzmann, Mathieu and Hu, Yinlin},
	title = {Rigidity-Aware Detection for {6D} Object Pose Estimation},
	booktitle = {CVPR},
	pages = {8927--8936},
	year = 2023
}

@inproceedings{wu_kcv_2022,
	author = {Wu, Yangzheng and Javaheri, Alireza and Zand, Mohsen and Greenspan, Michael},
	title = {Keypoint Cascade Voting for Point Cloud Based {6DoF} Pose Estimation},
	booktitle = {3DV},
	pages = {176--186},
	year = 2022
}

@inproceedings{haugaard_surfemb_2022,
	author = {Haugaard, Rasmus Laurvig and Buch, Anders Glent},
	title = {{SurfEmb}: Dense and Continuous Correspondence Distributions for Object Pose Estimation with Learnt Surface Embeddings},
	booktitle = {CVPR},
	pages = {6739--6748},
	year = 2022
}

@inproceedings{moon_genflow_2024,
	author = {Moon, Sungphill and Son, Hyeontae and Hur, Dongcheol and Kim, Sangwook},
	title = {{GenFlow}: Generalizable Recurrent Flow for {6D} Pose Refinement of Novel Objects},
	booktitle = {CVPR},
	pages = {10039--10049},
	year = 2024
}

@inproceedings{leonardis_foundpose_2025,
	author = {Örnek, Evin Pinar and Labbé, Yann and Tekin, Bugra and Ma, Lingni and Keskin, Cem and Forster, Christian and Hodan, Tomas},
	title = {FoundPose: Unseen Object Pose Estimation with Foundation Features},
	booktitle = {ECCV},
	pages = {163--182},
	year = 2024
}

@inproceedings{caraffa_freeze_2024,
	author = {Caraffa, Andrea and Boscaini, Davide and Hamza, Amir and Poiesi, Fabio},
	title = {{FreeZe}: Training-Free Zero-Shot {6D} Pose Estimation with Geometric and Vision Foundation Models},
	booktitle = {ECCV},
	pages = {414--431},
	year = 2024
}

@misc{freezev2,
	author = {Andrea Caraffa and Davide Boscaini and Fabio Poiesi},
	title = {Accurate and Efficient Zero-Shot {6D} Pose Estimation with Frozen Foundation Models},
	note = {ArXiv preprint arXiv:2506.09784},
	year = 2025
}

@inproceedings{CosyPose,
	author={Y. {Labbe} and J. {Carpentier} and M. {Aubry} and J. {Sivic}},
	title= {{CosyPose}: Consistent Multi-view Multi-object {6D} Pose Estimation},
	booktitle={ECCV},
	year={2020}}

@inproceedings{bib1,
	author={Kalra, Agastya and Stoppi, Guy and Marin, Dmitrii and Taamazyan, Vage and Shandilya, Aarrushi and Agarwal, Rishav and Boykov, Anton and Chong, Tze Hao and Stark, Michael},
	booktitle= {CVPR}, 
	title={Towards Co-Evaluation of Cameras, {HDR}, and Algorithms for Industrial-Grade {6DoF} Pose Estimation}, 
	year= 2024,
	pages={22691-22701}}

@inproceedings{bib2,
	author={Qian, Kun and Erden, Mustafa Suphi and Kong, Xianwen},
	booktitle= {IROS}, 
	title={Scalable Network and Adaptive Refinement Module for {6D} Pose Estimation of Diverse Industrial Components}, 
	year=2024,
	pages={11646-11653},
}

@inproceedings{bib3,
	title	= "{ConvPoseCNN}: Dense Convolutional {6D} Object Pose Estimation",
	author		= "Capellen, Catherine and Schwarz, Max and Behnke, Sven",
	booktitle= {VISIGRAPP},
	pages={162--172},
	year={2020}
}

@inproceedings{bib4,
	title={{CDPN}: Coordinates-Based Disentangled Pose Network for Real-Time {RGB}-Based {6-DoF} Object Pose Estimation},
	author={Li, Zhigang and Wang, Gu and Ji, Xiangyang},
	booktitle= {ICCV},
	pages={7678--7687},
	year={2019}
}

@inproceedings{bib5,
	title={{Pix2Pose}: Pixel-Wise Coordinate Regression of Objects for {6D} Pose Estimation},
	author={Park, Kiru and Patten, Timothy and Vincze, Markus},
	booktitle= {ICCV},
	pages={7668--7677},
	year={2019}
}

@article{epnp,
	title={{EPnP}: An Accurate {O}(n) Solution to the {PnP} Problem},
	author={Lepetit, Vincent and Moreno-Noguer, Francesc and Fua, Pascal},
	journal= {IJCV},
	volume= 81,
	pages={155--166},
	year=2009,
}

@inproceedings{nguyen2024gigaPose,
	title={{GigaPose}: Fast and Robust Novel Object Pose Estimation via One Correspondence},
	author={Nguyen, Van Nguyen and Groueix, Thibault and Salzmann, Mathieu and Lepetit, Vincent},
	booktitle = {CVPR},
	year={2024}}

@inproceedings{gso,
	title={Google Scanned Objects: A High-Quality Dataset of 3D Scanned Household Items},
	author = {Downs, Laura and Francis, Anthony and Koenig, Nate and Kinman, Brandon and Hickman, Ryan and Reymann, Krista and McHugh, Thomas B. and Vanhoucke, Vincent},
	booktitle= {ICRA},
	pages = {2553 - 2560},
	year = {2022},
}
